\documentclass{article}

\usepackage[preprint,nonatbib]{neurips_2022}

\usepackage[utf8]{inputenc} 
\usepackage[T1]{fontenc}    
\usepackage{hyperref}       
\usepackage{url}            
\usepackage{booktabs}       
\usepackage{amsfonts}       
\usepackage{nicefrac}       
\usepackage{microtype}      
\usepackage{xcolor}         

\usepackage{graphicx}
\usepackage{tikz}
\usepackage{amsmath,amssymb} 
\usepackage{color}

\usepackage{subcaption}
\usepackage{multirow}
\usepackage{makecell}
\usepackage{wrapfig}
\usepackage{enumerate}
\usepackage{algorithm}
\usepackage{bbm}
\usepackage{algpseudocode}

\newcommand{\eg}{\textit{e.g.}}
\newcommand{\ie}{\textit{i.e.}}

\newcommand{\etc}{\textit{etc}}

\title{Privacy-Preserving Model Upgrades with \\Bidirectional Compatible Training in Image Retrieval}

\author{
Shupeng Su$^1$\footnotemark[1] \qquad Binjie Zhang$^{1,2}$\footnotemark[1] \qquad Yixiao Ge$^1$\footnotemark[2] \\ \textbf{Xuyuan Xu$^3$} \qquad \textbf{Yexin Wang$^3$} \qquad \textbf{Chun Yuan$^2$} \qquad \textbf{Ying Shan$^1$} \\
~~\\
$^1$ARC Lab, Tencent PCG \quad $^2$Tsinghua University \quad
$^3$AI Technology Center of Tencent Video \\
\texttt{\{pennsu,yixiaoge\}@tencent.com} \quad \texttt{zbj19@mails.tsinghua.edu.cn}
}

\begin{document}

\maketitle
\renewcommand{\thefootnote}{\fnsymbol{footnote}}
\footnotetext[1]{Equal contribution. Work done during Binjie's internship at ARC Lab, Tencent PCG.}
\footnotetext[2]{Corresponding author.}

\begin{abstract}
The task of privacy-preserving model upgrades in image retrieval desires to reap the benefits of rapidly evolving new models without accessing the raw gallery images.
A pioneering work \cite{shen2020cvpr} introduced backward-compatible training, where the new model can be directly deployed in a backfill-free manner, \textit{i.e.}, the new query can be directly compared to the old gallery features.
Despite a possible solution, its improvement in sequential model upgrades is gradually limited by the fixed and under-quality old gallery embeddings. 
To this end, we propose a new model upgrade paradigm, termed Bidirectional Compatible Training (BiCT), which will upgrade the old gallery embeddings by forward-compatible training towards the embedding space of the backward-compatible new model.
We conduct comprehensive experiments to verify the prominent improvement by BiCT and interestingly observe that the inconspicuous loss weight of backward compatibility actually plays an essential role for both backward and forward retrieval performance.
To summarize, we introduce a new and valuable problem named privacy-preserving model upgrades, with a proper solution BiCT. Several intriguing insights are further proposed to get the most out of our method.

\end{abstract}

\section{Introduction}

Nowadays, image retrieval~\cite{alzu2015semantic,smoothap:brown2020smooth,liu2020scenesketcher,revaud2019learning,yu2018product,zhang2022towards,zhao2018modulation} receives growing concerns due to its extensive applications in industry.
Given a query image from client, the image retrieval system will return the relevant samples or the corresponding indices of gallery via the query-gallery similarities. While for the privacy-sensitive retrieval scenarios (\textit{e.g.}, person re-identification \cite{hermans2017defense,xiao2017joint,shen2018end,ge2020mutual}, face identification \cite{liu2017sphereface,wang2017normface,wang2018cosface,deng2019arcface}, copyright detection \cite{ke2004efficient,zhou2016effective,zhou2016effective2,liu2020content}), the retrieval systems are supposed to only retain the extracted gallery embeddings, discarding the raw images for privacy protection.
To harvest the benefits of rapidly evolving new models for these privacy-sensitive retrieval systems\footnote[3]{The conventional model upgrades in a retrieval system require to re-extract the gallery embeddings from raw images with the new model (dubbed as ``backfill'') before deployment.}, we investigate a new and valuable problem, namely \textbf{privacy-preserving model upgrades}, which aims to enable the model upgrades without accessing the raw gallery images.

\begin{figure}[t]
\centering
\includegraphics[width=0.9\linewidth]{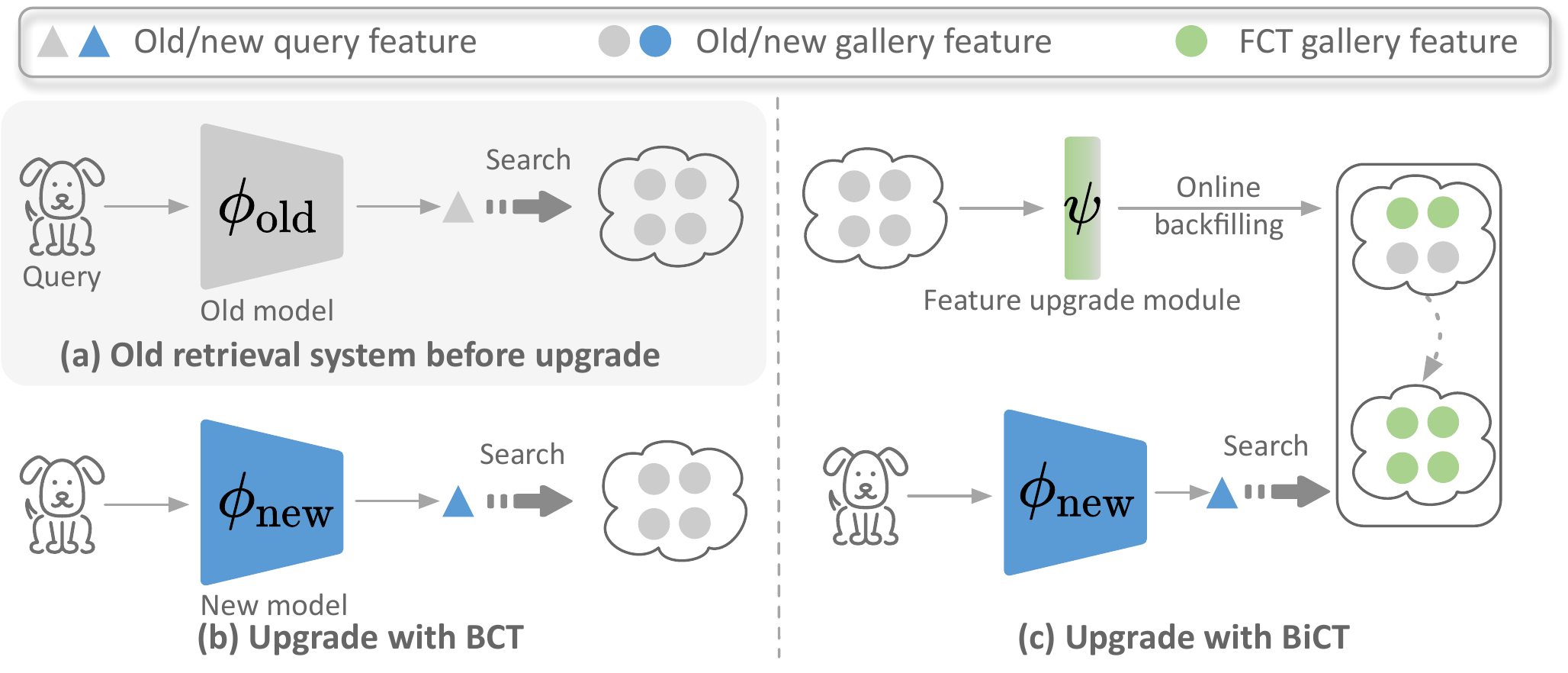}
\caption{
Given \textbf{(a)} an old retrieval system to be upgraded, we illustrate and compare the different model upgrading paradigms of \textbf{(b)} backward-compatible training (BCT) \cite{shen2020cvpr} and \textbf{(c)} our bidirectional compatible training (BiCT).}
\label{fig:teaser}
\end{figure}

To resolve that, the backward-compatible training (BCT) \cite{shen2020cvpr} is a workable solution which constrains the embeddings of new model to be compatible with the old ones.
Accordingly, the backward-compatible new model can be directly deployed on system without backfilling the gallery embeddings (dubbed as ``backfill-free'') and the system accuracy increases from the improved query features.
\begin{wraptable}{r}{6.4cm}
\renewcommand\arraystretch{1.2}
\footnotesize
\centering
\caption{The retrieval accuracy of sequential model upgrades with increasing training data, 25\%$\rightarrow$50\%$\rightarrow$100\%.
Details are shown later.}
\vspace{-0.2cm}
\setlength{\tabcolsep}{0.4mm}{	
\begin{tabular}{cccccc}
	\toprule
    \multirow{2}{*}{Method} & \multirow{2}{*}{${\cal M}_{\rm o2o}$} & \multicolumn{2}{c}{1st upgrade} & \multicolumn{2}{c}{2nd upgrade} \\
	\cmidrule(r){3-4}
	\cmidrule(r){5-6}
	&& ${\cal M}_{\rm BCT}$ & ${\cal M}_{\rm FCT}$ & ${\cal M}_{\rm BCT}$ & ${\cal M}_{\rm FCT}$ \\
	\midrule
	BCT \cite{shen2020cvpr} & \multirow{2}{*}{21.63} & 22.23 & n$\!/\!$a & 23.19 & n$\!/\!$a \\
	our BiCT  & &  22.23 & 23.60 & 24.38 & 25.53 \\
	\bottomrule
\end{tabular}
}
\label{tab:intro}
\end{wraptable}
Despite the feasibility, the performance promotions by mere BCT are limited in sequential model upgrades as the new model has to heavily sacrifice the discriminativeness of the new embeddings to maintain the compatibility with the fixed and under-quality old gallery embeddings (see Table~\ref{tab:intro}).

Towards this end, we propose a new \textbf{Bi}directional \textbf{C}ompatible \textbf{T}raining (\textbf{BiCT}) framework 
which aims to forward upgrade the old gallery embeddings even without the raw gallery images.
Specifically, the proposed BiCT performs backward-compatible training (BCT) of a new embedding model to improve the query embeddings, followed by forward-compatible training (FCT) of a feature-to-feature upgrade module to update the gallery embeddings towards the superior embedding space of new model.
As illustrated in Fig.~\ref{fig:teaser}, in our BiCT model upgrade framework, the BCT new model can be deployed on system immediately after training and the FCT feature upgrade module can progressively upgrade the old gallery embeddings in system on-the-fly\footnote[4]{Due to the compatibility of embedding spaces among the new model, old model and feature upgrade module, the upgraded gallery features can replace the old ones in system gradually on-the-fly.}.
We highlight that the \textit{bidirectional} compatible training properly 
relieves the new model's shackle of keeping compatible with the oldest gallery embeddings in sequential upgrades.
The notable performance boost shown in Table \ref{tab:intro} exactly demonstrates the efficacy of BiCT than the mere BCT mechanism.

One may ask, why not use FCT solely for model upgrades? We would like to point out three advantages of BiCT than mere FCT as follows.
(1)~Training new model with BCT constraint sometimes boost the retrieval performance of FCT, especially when the feature upgrade module is lightweight and difficult to update the old features to a completely incompatible new space (we show this with experiments later).
(2)~Inheriting the backfill-free benefit of BCT, the new model in BiCT can be deployed on system immediately after training
while if using FCT independently, the new model can only be deployed after the gallery features are totally refreshed .
(3)~Besides, the update of gallery embeddings in BiCT can be conducted progressively on-the-fly, \ie, the system performance gradually rises as the features upgrading and backfilling proceed, which is definitely unavailable by mere FCT.

To fully explore the potential of our proposed BiCT,
we conduct comprehensive pilot experiments on the possible factors that affect the cross-model retrieval performance, 
including the network capacities, the compatible formulations, the relevant compatible hyper-parameters, \etc.
We interestingly observe that the backward-compatible loss weight, $\lambda$, is actually a key factor which is overlooked by previous compatible training work. 
To fill the vacancy, we provide a thorough analysis for this parameter under the various model upgrade settings and we figure out that BCT and FCT can complement each other to achieve peak performance with a proper $\lambda$.

In a nutshell, our contributions are three-fold.
(1)~We introduce a new problem, namely privacy-preserving model upgrades, which is nontrivial for the real-world image retrieval systems. 
(2)~To tackle the problem, we provides a new model upgrading paradigm called Bidirectional Compatible Training (BiCT), with which the new model can be deployed on system immediately after training followed by the progressive upgrade of gallery embeddings on-the-fly.
(3)~We demonstrate that the backward-compatible loss weight actually plays an essential role in compatible training which provides intriguing insights for tuning the peak BCT and FCT performances when applying our proposed BiCT framework.

\section{Related Work}


\paragraph{Backward Compatible Training (BCT).}
\cite{shen2020cvpr} is the first work to introduce BCT by adding an influence loss in the training objective of new model which enforces the new features to approach the corresponding old class centroids. 
Later, \cite{budnik2020asymmetric} proposes the asymmetric retrieval task which is a similar scenario to BCT in that the gallery and query features are extracted by different embedding models. It studies the adaptation of the common metric learning losses into asymmetric retrieval constraints when training the new model, which finally introduces a variant combining the contrastive loss and regression loss.
\cite{meng2021learning} also study the BCT task and focuses on the design of compatibility loss. It proposes an alignment loss to align the new classifier weights with the old ones and a boundary loss for a tighter constraint of backward compatibility. Although making progress, the mere BCT will achieve declining gains in the sequential model upgrades as the new model has to maintain the compatibility to the oldest gallery embeddings. Accordingly, upgrading the gallery without raw images is called for in privacy-preserving retrieval systems.

\paragraph{Cross Model Compatibility (CMC).}
CMC training \cite{chen2019r3,wang2020unified} also targets the compatible retrieval among different embedding models. 
However, it concerns training transformation modules to map the features from different models to a common space.
Specifically, \cite{chen2019r3} first introduces the CMC problem in the face recognition occasion and tackle it with a R$^{3}$AN framework. R$^{3}$AN transforms the source embeddings into the target embeddings as a regression task which simultaneously reconstruts the original face image in the middle stage as training regularization. \cite{wang2020unified} further generalizes the CMC training in unlimited retrieval occasions and improve the CMC performance by proposing the RBT (Residual Bottleneck Transformation) module with a training scheme combining classification loss, regression loss and KL-divergence loss. \cite{meng2021learning} also applies their proposed compatibility loss for the CMC task and shows improvement likewise. 
Note that the CMC training is exactly like the variant of our BiCT framework which only uses the FCT for model upgrades. However, as analysed in the Introduction, the mere FCT will lose the merits of the immediate deployment of new embedding model and progressively upgrading the gallery features on-the-fly, which even depresses the FCT performance in lightweight feature transformation module as shown later.

\paragraph{Hot-refresh model upgrades.}
Based on the BCT technique, \cite{zhang2022iclr} further introduces the hot-refresh model upgrades which backfills the gallery on-the-fly with the new embeddings extracted by the backward-compatible new model. It observes a model regression problem in the hot-refresh procedure and alleviates it by adding an explicit constraint for the new-to-new negative pairs in the backward-compatible training of new model as well as an uncertainty-based backfilling strategy for gallery refresh. 
However, the proposed hot-refresh mechanism needs to access the raw gallery images while in this paper we further generalize it for the privacy-preserving occasion with the proposed forward-compatible training of a feature upgrade module.

\section{Methodology}

In this paper, we investigate a new and practical problem dubbed privacy-preserving model upgrades which desires to upgrade the embedding model in retrieval system without accessing the raw gallery images (Sec. \ref{sec:prob}).
A previous work, backward-compatible training \cite{shen2018end}, provides a possible solution (Sec. \ref{sec:bct}) but shows limited performance gains in sequential model upgrades due to the fixed gallery embeddings.
Thus, we propose a new model upgrade paradigm, bidirectional compatible training (Sec. \ref{sec:bict}), in which the forward-compatible training is introduced to update the gallery embeddings after the deployment of the backward-compatible new model.
We elaborate each part as follows.

\subsection{Privacy-Preserving Model Upgrades in Image Retrieval}\label{sec:prob}
We denote the gallery image set as $\mathcal{G}$, and the query as $\mathcal{Q}$. 
With an embedding model $\phi: \text{Img} \mapsto \mathbb{R}^{d}$, images are mapped into compact feature vectors with a dimension of $d$. 
Given query $\mathcal{Q}$, image retrieval system aims to identify the object/content of interest in gallery resorting to the query-gallery feature similarities, \eg, the Euclidean distance $\|\phi(\mathcal{Q})-\phi(\mathcal{G})\|_{2}$ or cosine similarity $\cos(\phi(\mathcal{Q}), \phi(\mathcal{G}))$.
Upgrading the embedding model to achieve more discriminative image embeddings is a straightforward way to improve the accuracy of a retrieval system.
However, the conventional model upgrades need to re-extract the gallery embeddings (termed as ``backfilling'') with the new model before uploading it to replace the old model.
Such a practice is infeasible when the raw gallery images cannot be retained in system due to the common privacy issues, \eg, images from the surveillance scenarios or copyrighted data.
We accordingly introduce a new problem, namely privacy-preserving model upgrades, to reap the benefit of rapidly evolving new models without accessing the raw gallery images. It is an essential but intractable problem for the real-world image retrieval systems.

\subsection{Backward Compatible Training (BCT): A Revisit}\label{sec:bct}
To tackle above problem, a previous work, the backward-compatible training (BCT)~\cite{shen2020cvpr} can serve as a workable solution which constrains the embeddings of the new model to be compatible with the ones from old model. 
It introduces a backfill-free paradigm for model upgrades, that is, the new model can be directly deployed on system without refreshing the gallery features.
The retrieval accuracy of the new system is improved by the advanced query features from new model.
Accordingly, the objective of BCT in sequential model upgrades can be formulated as
\begin{equation}
\label{eq:ppmu-inequality}
    {\cal M}(\phi_{\rm old}(\mathcal{Q}), \phi_{\rm old}(\mathcal{G})) <
    {\cal M}(\phi_{\rm new}^{\rm 1st}(\mathcal{Q}), \phi_{\rm old}(\mathcal{G})) < 
    {\cal M}(\phi_{\rm new}^{\rm 2nd}(\mathcal{Q}), \phi_{\rm old}(\mathcal{G})),
\end{equation}
where ${\cal M}(\cdot,\cdot)$ denotes the evaluation metric for retrieval (\textit{e.g.}, mAP), and we demonstrate two generations of model upgrades here.
Despite the feasibility, the improvements by BCT are highly limited especially in later upgrade generations as the new model has to sacrifice the discriminativeness to maintain compatibility with the oldest gallery, rendering BCT a sub-optimal solution to tackle the challenge of privacy-preserving model upgrades.

\subsection{Bidirectional Compatible Training (BiCT)}\label{sec:bict}

Therefore, to harvest more benefits from the new model, the gallery embeddings are also supposed to be upgraded in a way without accessing the raw gallery images.
Towards this end, we introduce a Bidirectional Compatible Training (BiCT) framework, which performs backward-compatible training of a new model to improve the query embeddings, followed by forward-compatible training of a feature upgrade module to update the gallery embeddings. We formally introduce it as follows.

\begin{figure}[t]
\centering
\includegraphics[width=0.95\linewidth]{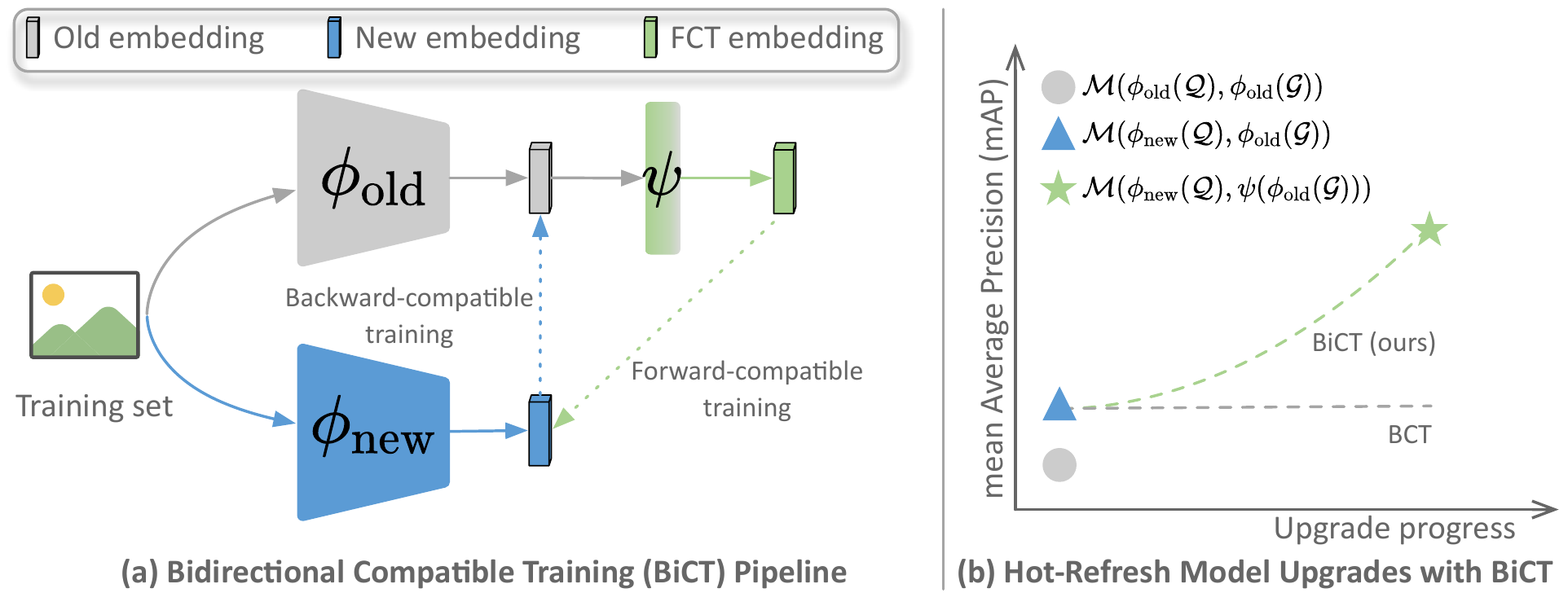}
\caption{(a) The proposed {Bidirectional Compatible Training (BiCT)} framework which performs the forward-compatible training of a feature upgrade module after the backward-compatible training of new embedding model. (b) The mAP trend in a model upgrade with our BiCT framework which shows a step performance growth from BCT followed by a gradual increase with FCT.}
\label{fig:framework}
\end{figure}

\paragraph{Architecture of BiCT.}
As shown in Fig.~\ref{fig:framework}, our BiCT framework consists of a new embedding model $\phi_{\rm new}$ (\textit{e.g.}, CNNs~\cite{he2016deep,ge2021self,ge2020mutual} or ViTs~\cite{liu2021swin,ge2022bridgeformer}) and a feature-to-feature upgrade module $\psi$ (\textit{e.g.}, MLPs). 
The new embedding model is trained with backward compatibility to the embeddings of old model $\phi_{\rm old}$ and the feature-to-feature upgrade module is trained with forward compatibility to the embeddings of new model.
Thus we also dub the new embedding model as BCT model, and the feature-to-feature upgrade module as FCT module.
The new query features produced by $\phi_{\rm new}$ can directly probe among the old gallery, which is denoted as $\phi_{\rm new}(\mathcal{Q}) \to \phi_{\rm old}(\mathcal{G})$.
And $\psi$ updates $\phi_{\rm old}(\mathcal{G})$ subsequently towards the latent space of $\phi_{\rm new}$ to achieve better retrieval results, \textit{i.e.}, $\phi_{\rm new}(\mathcal{Q}) \to \psi(\phi_{\rm old}(\mathcal{G}))$.

\paragraph{Model upgrades with BiCT.}

Accordingly, the objective of BiCT in sequential model upgrades is formulated as follows
\begin{small} 
\begin{align}
\label{eq:ppmu-inequality-2}
    &{\cal M}(\phi_{\rm old}(\mathcal{Q}), \phi_{\rm old}(\mathcal{G})) <
    {\cal M}_\text{BCT}(\phi_{\rm new}^{\rm 1st}(\mathcal{Q}), \phi_{\rm old}(\mathcal{G})) < 
    {\cal M}_\text{FCT}(\phi_{\rm new}^{\rm 1st}(\mathcal{Q}), \psi^{\rm 1st}(\phi_{\rm old}(\mathcal{G}))) \nonumber \\
    &<{\cal M}_\text{BCT}(\phi_{\rm new}^{\rm 2nd}(\mathcal{Q}), \psi^{\rm 1st}(\phi_{\rm old}(\mathcal{G})))<
    {\cal M}_\text{FCT}(\phi_{\rm new}^{\rm 2nd}(\mathcal{Q}), \psi^{\rm 2nd}(\psi^{\rm 1st}(\phi_{\rm old}(\mathcal{G})))),    
\end{align}
\end{small}
where ${\cal M}_\text{BCT}$ indicates the BCT performance when the new model is deployed on system immediately after training, ${\cal M}_\text{FCT}$ indicates the FCT performance after the backfilling of the upgraded gallery embeddings. Compared with the objective of mere BCT in Equation~\eqref{eq:ppmu-inequality}, BiCT will achieve more and more prominent performance increase than mere BCT in the sequential model upgrades due to the up-to-date gallery embeddings.

\begin{algorithm}[t]
	\caption{Pipeline of Bidirectional Compatible Training.}
	\textbf{Input:} Training set ${\cal D}$, Old model $\phi_{\rm old}$
	
	\textbf{Step1:} Fix the parameters of $\phi_{\rm old}$ 
	\begin{algorithmic}[1]
		\While {not converges}
		\State update $\phi_{\rm new}$ by minimizing ${\cal L}_{\rm BCT}$ in Equation (\ref{eq:bct-loss})
		\EndWhile
	\end{algorithmic} 
	
	\textbf{Step2:} Fix the parameters of $\phi_{\rm old}$ and $\phi_{\rm new}$  
	\begin{algorithmic}[1]
		\While {not converges}
		\State update $\psi$ by minimizing ${\cal L}_{\rm FCT}$ in Equation (\ref{eq:fct-one-loss})
		\EndWhile
	\end{algorithmic} 
	
	\textbf{Output:} $\phi_{\rm new}$, $\psi$ 
	\label{alg:train}
\end{algorithm}



\paragraph{Training objectives of BiCT.}
The overall training objective of BiCT can be formulated as
\begin{align}
    \begin{split}
        {\psi}=\mathop{\arg\min}_{\psi}~ {\mathcal{L}_\text{FCT}},
    ~~~~\text{s.t.}~~ \phi_{\rm new}= \mathop{\arg\min}_{\phi_\text{new}}~ {\mathcal{L}_\text{BCT}},
    \end{split}
\end{align}
where the new model $\phi_{\rm new}$ is first optimized with backward-compatible regularizations, and the feature upgrade module $\psi$ is subsequently learned with forward-compatible constraint.
The training pipeline is specified in Alg.~\ref{alg:train}.

Detailedly, for the backward-compatible training of new model $\phi_\text{new}$, two objectives are taken into consideration: (1) obtaining discriminative new representations, and (2) being compatible to the old embedding space. Thus, the objective is formulated as
%
\begin{equation} \label{eq:bct-loss}
  \mathcal{L}_\text{BCT}(\phi_{\rm new}) = \mathcal{L}_{\rm base}(\phi_{\rm new}) + \lambda\ell_{\rm comp}(\phi_{\rm new}, \phi_{\rm old}),
\end{equation}
where $\mathcal{L}_{\rm base}(\phi_{\rm new})$ is the basic feature discriminativeness loss for the new model and $\ell_{\rm comp}(\phi_{\rm new}, \phi_{\rm old})$ is the compatible constraint for $\phi_{\rm new}$ towards $\phi_{\rm old}$.
$\lambda$ is the loss weight coefficient for backward compatibility.
Specifically, following state-of-the-art metric learning methods~\cite{gldv2-1rd:jeon20201st,gldv2-3rd:mei20203rd} in image retrieval, we adopt ArcFace loss~\cite{deng2019arcface} as the basic loss ($\mathcal{L}_{\rm base}=\ell_{\rm arc}(\phi_{\rm new},\omega_{\rm new})$), that is,
\begin{equation}\label{eq:arcface_loss}
  \begin{split}
      \ell_{\rm arc}(\phi,\omega)
      = -\frac{1}{|{\cal D}|}\sum_{i\in {\cal D}}{ \log{ \frac{e^{s\cdot k(\phi(i),\omega(i), m)}}{e^{s\cdot k(\phi(i),\omega(i), m)}+\sum_{j\neq y_i}e^{s\cdot k(\phi(j),\omega(i), 0)} }}}, 
  \end{split}
\end{equation}
where $m$ is the margin and $s$ is the scaler hyper-parameters. $\mathcal{D}$ denotes the training image set and $y_{i}$ denotes the label of image $i$ in ${\cal D}$. 
The kernel function is defined as $k(\phi(i),\omega(i), m)=\cos(\arccos\langle\phi(i),\omega(i)\rangle+m)$, and $\langle\cdot,\cdot\rangle$ represents the inner product.
%
The compatible loss ($\ell_{\rm comp}$) constrains the new embeddings to approach the corresponding old features, which can be formulated as a regression loss,
\begin{equation} \label{eq:cosine-embedding-loss}
  \ell_{\rm comp}(\phi_{\rm new}, \phi_{\rm old}) = \frac{1}{|{\cal D}|}\sum_{i \in \mathcal{D}} [1 - \cos\langle\phi_{\rm new}(i), \phi_{\rm old}(i)\rangle].
\end{equation}

As for the forward-compatible training of feature upgrade module $\psi$, it only needs to concern the compatibility between the embeddings of $\psi$ towards $\phi_{\rm new}$, hence,
\begin{equation} \label{eq:fct-one-loss}
  \mathcal{L}_{\rm FCT}(\psi) = \ell_{\rm comp}( \phi_{\rm new}, \psi(\phi_{\rm old})).
\end{equation}

We would like to highlight again that in the model upgrade with BiCT, the module $\psi$ only operates on the old gallery embeddings instead of the raw images.
Thereby it properly resolves the dilemma between the absence of raw gallery images due to the privacy protection and the demand of upgrading the outdated gallery embeddings, no matter for the superior retrieval performance in the current model generation (\ie, $\mathcal{M}_{\rm FCT}^{\rm 1st}>\mathcal{M}_{\rm BCT}^{\rm 1st}$) or the future model upgrades (\ie, better $\mathcal{M}_{\rm BCT}^{>\rm 1st}$ of BiCT than mere BCT). The right part of Fig.~\ref{fig:framework} depicts the trend of retrieval performance during a  model upgrade experiment with our BiCT framework. It indeed shows a step performance increase from the immediate deployment of BCT model $\phi_{\rm new}$, followed by a gradual increase with the refresh of $\phi_{\rm old}(\mathcal{G})$ using the FCT module $\psi$.

\paragraph{Analysis of BiCT.}
\label{sec:framework_analysis}
\begin{wrapfigure}{r}{0.5\textwidth}
  \vspace{-30pt}
  \begin{center}
    \includegraphics[width=0.45\textwidth]{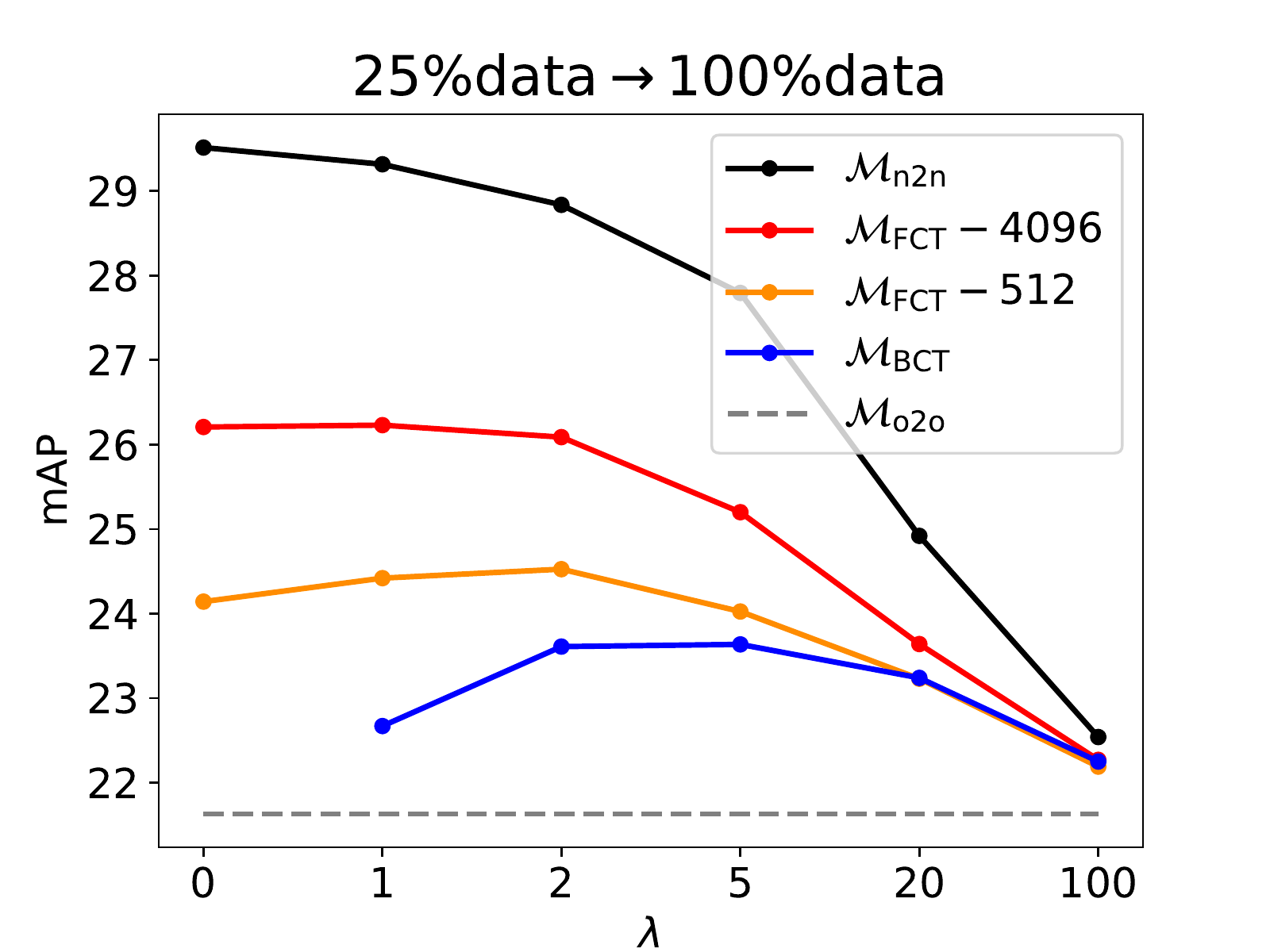}
    \vspace{-2pt}
    \caption{Retrieval performance of BiCT under different values of $\lambda$ and hidden dimensions of FCT module $\psi$ when model upgrades from $25\%$ training data to $100\%$. Performance notations are indicated in Table~\ref{tab:metric}.
    }
    \label{fig:lambda-data-alone}
  \end{center}
  \vspace{-20pt}
\end{wrapfigure}

The backward-compatible loss weight (\ie, $\lambda$ in Equation~\eqref{eq:bct-loss}) controls the balance between the new feature discriminativeness and the backward compatibility. It is proven to be a critical factor for the cross-model retrieval performance in our study but overlooked by the previous work~\cite{shen2020cvpr,meng2021learning,zhang2022iclr} (detailed experiments and discussions are provided in Section~\ref{sec:lambda-analysis}).
The retrieval performances with varying $\lambda$ under our BiCT framework are exhibited in Fig.~\ref{fig:lambda-data-alone},
from which we provide two intriguing insights as follows:

\noindent (1) As will be demonstrated in Section~\ref{sec:lambda-analysis}, ${\cal M}_{\rm BCT}$ (blue curve in Fig.~\ref{fig:lambda-data-alone}) will always present a peak trend (we denote the peak position as $\lambda_{b}$).
Besides, with a lightweight $\psi$ (orange curve) ${\cal M}_{\rm FCT}$ also constitutes peak (denoted as $\lambda_{f}$),
but with a burdensome $\psi$ (red curve) ${\cal M}_{\rm FCT}$ decreases slowly as first and accelerate later.
Based on those, a retrieval system with BiCT should pay attention to $\lambda_{b}$ if desiring the best backward retrieval performance. While setting $\lambda$ between 0 and $\lambda_{b}$ can achieve high forward retrieval performance, a decent compromise for both ${\cal M}_{\rm BCT}$ and ${\cal M}_{\rm FCT}$ lies between $\lambda_{f}$ and $\lambda_{b}$. 

\noindent (2)
Performing BCT ($\lambda>0$) for $\phi_{\rm new}$ will benefit ${\cal M}_{\rm FCT}$ if the feature upgrade module $\psi$ is lightweight (the beginning of orange curve), as the more compatible new model will mitigate the difficulty for $\psi$ to upgrade the old embeddings towards the new ones.
In turn, performing FCT after BCT can bring consistent improvements for the retrieval performance (\ie, ${\cal M}_{\rm FCT}>{\cal M}_{\rm BCT}$) which will benefit the BCT in next upgrade generation as it relieves the shackle of next new model for keeping compatibility to the oldest and under-quality gallery embeddings.
Therefore, BCT and FCT are complementary to each other which is exactly the essential merit of our proposed Bidirectional Compatible Training in sequential model upgrades. 


\section{Experiments}
In this section, we validate the proposed BiCT on the widely-acknowledged image retrieval datasets, including the Google Landmark v2~\cite{gldv2_2020}, $\mathcal{R}$Oxford~\cite{roxford_rparis_2018} and $\mathcal{R}$Paris~\cite{roxford_rparis_2018}.
The relevant codes will be public soon.

\subsection{Experiment settings}
\paragraph{Datasets.} 
We adopt the GLDv2-train-clean version~\cite{gldv2_2020} as training set which is comprised of 1,580,470 images in 81,313 landmarks. To simulate the different model upgrade occasions, we randomly split the training set with different percentages (25\%,50\%, and 100\%) which are specified in detail in Table~\ref{tab:dataset}. For the extended-data scenario (\eg, 25\%data $\rightarrow$ 100\%data), the old training set and the new one share all landmarks. While for the extended-class scenario (\eg, 25\%class $\rightarrow$ 100\%class), the old training set only covers a part.
After the training, the retrieval performance is evaluated on GLDv2-test~\cite{gldv2_2020}, Revisited Oxford ($\mathcal{R}$Oxford)~\cite{roxford_rparis_2018}, and Revisited Paris ($\mathcal{R}$Paris)~\cite{roxford_rparis_2018}. 
GLDv2-test has 750 query images and 761,757 gallery images. $\mathcal{R}$Oxford contains 70 query images and 4,993 gallery images while $\mathcal{R}$Paris has 70 query images likewise and 6,322 gallery images.

\begin{figure}[t]
 \begin{minipage}[t]{0.47\textwidth}
  \centering
  \makeatletter\def\@captype{table}\makeatother\caption{Different allocations for the training data, where all the images are sampled from GLDv2-train-clean.}
        \centering
        \setlength{\tabcolsep}{4 pt}
        \begin{tabular}{lccc}
            \toprule
            ~ & Usage &  Images &  Classes \\
            \midrule
            \multirow{3}{*}{\makecell[c]{ {Data-split} }} & 25\% & 374,264 & 81,313 \\
            ~ & 50\% & 771,137 & 81,313 \\
            ~ & 100\% & 1,580,470 & 81,313\\
            \cmidrule(r){1-4}
            \multirow{3}{*}{\makecell[c]{ {Class-split} }} & 25\% & 389,526 & 20,328 \\
            ~ & 50\% &  791,049 & 40,656 \\
            ~ & 100\% &  1,580,470 & 81,313 \\
            \bottomrule
        \end{tabular}
        \label{tab:dataset}
  \end{minipage}
  \hfill
  \begin{minipage}[t]{0.49\textwidth}
   \centering
    \makeatletter\def\@captype{table}\makeatother\caption{Performance notations in one generation model upgrade. ${\cal M}_{\rm n2n}$ is an oracle and cannot reach without the raw gallery images.}
        \centering
        \setlength{\tabcolsep}{10 pt}{
        \begin{tabular}{ccc}
            \toprule
            \multirow{2}{*}{\makecell[c]{Symbol}} & \multicolumn{2}{c}{Embedding} \\
            \cmidrule(r){2-3}
            ~ & Query & Gallery \\
            \midrule
            ${\cal M}_{\rm o2o}$ & $\phi_{\rm old}(\mathcal{Q})$ & $\phi_{\rm old}(\mathcal{G})$ \\
            ${\cal M}_{\rm BCT}$ & $\phi_{\rm new}(\mathcal{Q})$ & $\phi_{\rm old}(\mathcal{G})$\\
            ${\cal M}_{\rm FCT}$ & $\phi_{\rm new}(\mathcal{Q})$ & $\psi(\phi_{\rm old}(\mathcal{G}))$ \\
            ${\cal M}_{\rm n2n}$ & $\phi_{\rm new}(\mathcal{Q})$ & $\phi_{\rm new}(\mathcal{G})$ \\
            \bottomrule
        \end{tabular}
        }
        \label{tab:metric}
   \end{minipage}
\end{figure}

\paragraph{Metric.}
We adopt the cosine similarity for ranking and the retrieval results are measured with the mean Average Precision (mAP). 
Specifically, following the test protocols proposed in \cite{gldv2_2020,roxford_rparis_2018}, we use mAP@100 for GLDv2-test, and mAP in the \textbf{Medium} evaluation setup (\textit{Easy} and \textit{Hard} images are treated as positive, while \textit{Unclear} are ignored) for $\mathcal{R}$Oxford and $\mathcal{R}$Paris.
The performance notations are listed in Table \ref{tab:metric}.

\paragraph{Training details.}
Most of our basic settings follows the experience in \cite{yuqi20212nd,ozaki2019large,henkel2020supporting}. Specifically, we adopt the ResNet~\cite{he2016deep} (resnet50 by default unless specified otherwise) as the backbone of embedding model and substitute the global average pooling layer by Generalized-Mean (GeM) pooling~\cite{radenovic2018fine} with hyper-parameter p=3. A fully-connected (fc) layer is appended in the end which transforms the pooling features into the output embedding (the dimension is set as 512). 
For the feature upgrade module $\psi$, we adopt [fc-bn-relu] as basic block and stack 3 of this with a output fc layer as the final MLP construction. 
The input image is resized to $384\times 384$ for training and $640\times 640$ for inference. 
Random image augmentation is applied which includes the random resized cropping and horizontal flipping.
SGD optimizer with 0.9 momentum and 10$^{-4}$ weight decay is adopted. Besides, we uniformly use the cosine lr scheduler with 1 warm-up epoch in the total running of 30 epochs. The initial learning rate is set as 0.1 for the embedding model and 1 for MLP.
For the training objective, the ArcFace loss is adopted with hyper-parameter $s=30$ and $m=0.3$.

\subsection{Privacy-preserving model upgrades with BiCT}

\paragraph{Different hidden dimensions of MLP.}
We first ablate the hidden dimension $d_{h}$ of the feature upgrade module $\psi$. Results in Table~\ref{tab:exp-diff-fct} indicate that the ${\cal M}_{\rm FCT}$ grows with the increase of $d_{h}$. 
In addition, FCT consistently brings further performance boost after BCT in all evaluation datasets,  which exactly demonstrates the potential and necessity of performing forward-compatible training to upgrade the old gallery embeddings.
Considering the trade-off between the accuracy and the refresh efficiency, we adopt the median hidden dimension 4096 as the default setting in the following experiments unless stated otherwise.

\begin{table}[tb]
\renewcommand\arraystretch{1.2}
\centering
\caption{Ablation of the hidden dimension in MLP as $\psi$ for FCT performance (${\cal M}_{\rm FCT}$) when model upgrades with 25\% data $\rightarrow$ 100\% with BiCT ($\lambda=2$).
The numbers in \textcolor[RGB]{0,0,200}{blue} are the improvements than ${\cal M}_{\rm BCT}$.}
\vspace{0.1cm}
\setlength{\tabcolsep}{0.8mm}{	
\begin{tabular}{lcclllllc}
	\toprule
	\multirow{2}[3]{*}{Dataset} & \multirow{2}[3]{*}{${\cal M}_{\rm o2o}$} & \multirow{2}[3]{*}{${\cal M}_{\rm BCT}$} & \multicolumn{5}{c}{${\cal M}_{\rm FCT}$} & \multirow{2}[3]{*}{\textcolor[RGB]{150,150,150}{${\cal M}_{\rm n2n}$}}
	\\
	\cmidrule(r){4-8}
	&&& 512 & 1024 & 4096 & 8192 & 16384 &
	\\
	\midrule
	GLDv2 & 21.63 
	& 23.61 
	& 24.53 {\textcolor[RGB]{0,0,200}{\scriptsize $\!\!\uparrow\!0.92$}}
	& 24.90 {\textcolor[RGB]{0,0,200}{\scriptsize $\!\!\uparrow\!1.29$}}
	& 26.09 {\textcolor[RGB]{0,0,200}{\scriptsize $\!\!\uparrow\!2.48$}}
	& 26.58 {\textcolor[RGB]{0,0,200}{\scriptsize $\!\!\uparrow\!2.97$}}
	& 26.86 {\textcolor[RGB]{0,0,200}{\scriptsize $\!\!\uparrow\!3.25$}}
	& \textcolor[RGB]{150,150,150}{28.84}
	\\
	$\mathcal{R}$Oxford & 67.76 
	& 69.76
	& 70.84 {\textcolor[RGB]{0,0,200}{\scriptsize $\!\!\uparrow\!1.08$}}
	& 71.21 {\textcolor[RGB]{0,0,200}{\scriptsize $\!\!\uparrow\!1.45$}}
	& 72.20 {\textcolor[RGB]{0,0,200}{\scriptsize $\!\!\uparrow\!2.44$}}
	& 72.43 {\textcolor[RGB]{0,0,200}{\scriptsize $\!\!\uparrow\!2.67$}}
	& 73.33 {\textcolor[RGB]{0,0,200}{\scriptsize $\!\!\uparrow\!3.57$}}
	& \textcolor[RGB]{150,150,150}{77.99}
	\\
	$\mathcal{R}$Paris & 83.95
	& 85.45
	& 86.30 {\textcolor[RGB]{0,0,200}{\scriptsize $\!\!\uparrow\!0.85$}}
	& 86.56 {\textcolor[RGB]{0,0,200}{\scriptsize $\!\!\uparrow\!1.11$}}
	& 86.94 {\textcolor[RGB]{0,0,200}{\scriptsize $\!\!\uparrow\!1.49$}}
	& 87.07 {\textcolor[RGB]{0,0,200}{\scriptsize $\!\!\uparrow\!1.62$}}
	& 87.18 {\textcolor[RGB]{0,0,200}{\scriptsize $\!\!\uparrow\!1.73$}}
	& \textcolor[RGB]{150,150,150}{88.82}
	\\
	\bottomrule
\end{tabular}
}
\label{tab:exp-diff-fct}
\end{table}

\begin{table}[t]
\renewcommand\arraystretch{1.2}
\centering
\caption{Model upgrade with BiCT ($\lambda=2$) in four common upgrade occasions: extended data, extended class, improved architecture and improved training objective. 
The numbers in \textcolor[RGB]{0,200,0}{green} are the improvements than ${\cal M}_{\rm o2o}$ and the numbers in \textcolor[RGB]{200,0,0}{red} are the declines.}
\vspace{0.1cm}
\setlength{\tabcolsep}{0.1mm}{	
\begin{tabular}{lcccccccccccc}
	\toprule
	\multirow{2}[3]{*}{Setting} & \multicolumn{4}{c}{GLDv2} & \multicolumn{4}{c}{$\mathcal{R}$Oxford} &
	\multicolumn{4}{c}{$\mathcal{R}$Paris} \\
	\cmidrule(r){2-5}
	\cmidrule(r){6-9}
	\cmidrule(r){10-13}
	& ${\cal M}_{\rm o2o}$ & ${\cal M}_{\rm BCT}$ & ${\cal M}_{\rm FCT}$ & \textcolor[RGB]{150,150,150}{${\cal M}_{\rm n2n}$}
	& ${\cal M}_{\rm o2o}$ & ${\cal M}_{\rm BCT}$ & ${\cal M}_{\rm FCT}$ & \textcolor[RGB]{150,150,150}{${\cal M}_{\rm n2n}$}
	& ${\cal M}_{\rm o2o}$ & ${\cal M}_{\rm BCT}$ & ${\cal M}_{\rm FCT}$ & \textcolor[RGB]{150,150,150}{${\cal M}_{\rm n2n}$} \\
	\midrule

	\multirow{2}{*}{\makecell[l]{25\%data\\$\rightarrow$100\%data}} &
	\multirow{2}{*}{21.63} & 23.61 & 26.09 & \multirow{2}{*}{\textcolor[RGB]{150,150,150}{28.84}} 
	& \multirow{2}{*}{67.76} & 69.76 & 72.20 & \multirow{2}{*}{\textcolor[RGB]{150,150,150}{77.99}} 
	& \multirow{2}{*}{83.95} & 85.45 & 86.94 & \multirow{2}{*}{\textcolor[RGB]{150,150,150}{88.82}} 
	\\
	&& {\textcolor[RGB]{0,200,0}{$\uparrow\!1.98$}} & {\textcolor[RGB]{0,200,0}{$\uparrow\!4.46$}}
	&&& {\textcolor[RGB]{0,200,0}{$\uparrow\!2.00$}} & {\textcolor[RGB]{0,200,0}{$\uparrow\!4.44$}}
	&&& {\textcolor[RGB]{0,200,0}{$\uparrow\!1.50$}} & {\textcolor[RGB]{0,200,0}{$\uparrow\!2.99$}}
	\\
	\hline

	\multirow{2}{*}{\makecell[l]{25\%class\\$\rightarrow$100\%class}} &
	\multirow{2}{*}{21.48} & 23.08 & 25.71 & \multirow{2}{*}{\textcolor[RGB]{150,150,150}{28.95}} 
	& \multirow{2}{*}{69.74} & 69.62 & 72.37 & \multirow{2}{*}{\textcolor[RGB]{150,150,150}{78.58}} 
	& \multirow{2}{*}{82.53} & 83.77 & 85.74 & \multirow{2}{*}{\textcolor[RGB]{150,150,150}{88.50}} 
	\\
	&& {\textcolor[RGB]{0,200,0}{$\uparrow\!1.60$}} & {\textcolor[RGB]{0,200,0}{$\uparrow\!4.23$}}
	&&& {\textcolor[RGB]{200,0,0}{$\downarrow\!0.12$}} & {\textcolor[RGB]{0,200,0}{$\uparrow\!2.63$}}
	&&& {\textcolor[RGB]{0,200,0}{$\uparrow\!1.24$}} & {\textcolor[RGB]{0,200,0}{$\uparrow\!3.21$}}
	\\
	\hline

	\multirow{2}{*}{\makecell[l]{resnet18\\$\rightarrow$resnet50}} &
	\multirow{2}{*}{23.89} & 25.21 & 27.04 & \multirow{2}{*}{\textcolor[RGB]{150,150,150}{29.35}} 
	& \multirow{2}{*}{73.83} & 73.95 & 74.83 & \multirow{2}{*}{\textcolor[RGB]{150,150,150}{78.99}} 
	& \multirow{2}{*}{85.97} & 86.23 & 87.09 & \multirow{2}{*}{\textcolor[RGB]{150,150,150}{88.61}} 
	\\
	&& {\textcolor[RGB]{0,200,0}{$\uparrow\!1.32$}} & {\textcolor[RGB]{0,200,0}{$\uparrow\!3.15$}}
	&&& {\textcolor[RGB]{0,200,0}{$\uparrow\!0.12$}} & {\textcolor[RGB]{0,200,0}{$\uparrow\!1.00$}}
	&&& {\textcolor[RGB]{0,200,0}{$\uparrow\!0.26$}} & {\textcolor[RGB]{0,200,0}{$\uparrow\!1.12$}}
	\\
	\hline

	\multirow{2}{*}{\makecell[l]{softmax\\$\rightarrow$arcface}} &
	\multirow{2}{*}{18.26} & 22.86 & 26.69 & \multirow{2}{*}{\textcolor[RGB]{150,150,150}{29.05}} 
	& \multirow{2}{*}{64.05} & 68.26 & 72.69 & \multirow{2}{*}{\textcolor[RGB]{150,150,150}{77.87}} 
	& \multirow{2}{*}{83.19} & 85.85 & 87.85 & \multirow{2}{*}{\textcolor[RGB]{150,150,150}{89.40}} 
	\\
	&& {\textcolor[RGB]{0,200,0}{$\uparrow\!4.60$}} & {\textcolor[RGB]{0,200,0}{$\uparrow\!8.43$}}
	&&& {\textcolor[RGB]{0,200,0}{$\uparrow\!4.21$}} & {\textcolor[RGB]{0,200,0}{$\uparrow\!8.64$}}
	&&& {\textcolor[RGB]{0,200,0}{$\uparrow\!2.66$}} & {\textcolor[RGB]{0,200,0}{$\uparrow\!4.66$}}
	\\
	\bottomrule
\end{tabular}
}
\label{tab:exp-one-generation-upgrade}
\end{table}

\paragraph{Different model upgrade occasions with BiCT.}
Now we validate our proposed BiCT comprehensively on the common and practical model upgrade occasions, including two on the growing training set size: 25\%data $\rightarrow$ 100\%data and 25\%class $\rightarrow$ 100\%class, one on the improved model architecture: resnet18 $\rightarrow$ resnet50, and one on the improved training objective: softmax $\rightarrow$ arcface. The results are shown in Table~\ref{tab:exp-one-generation-upgrade}, from which we can observe that the BCT generally improves the retrieval performances (except the extended class upgrade in $\mathcal{R}$Oxford) while FCT facilitates the further performance boost consistently. We call out the two notable merits of our BiCT framework: (1) the retrieval system can enjoy the promotion by BCT immediately after the deployment of new embedding model; (2) the system gradually harvest the promotion by FCT as the gallery feature upgrading proceeds.

\subsection{Empirical Study on the Backward-compatible Loss Weight $\lambda$}
\label{sec:lambda-analysis}

We briefly exhibit the importance of the compatible loss weight $\lambda$ in Section~\ref{sec:bict} and here we would like to provide more results on this intriguing factor.

\paragraph{Different regularizers share similar tendencies.}

We first conduct experiments with several common compatible regularizers under different $\lambda$. Besides the cosine regression loss in Equation~(\ref{eq:cosine-embedding-loss}), we supplement the classification version \cite{shen2020cvpr} which constrains the new embedding to approach the corresponding old class center. Here we adopt the ArcFace loss again,
\begin{equation} \label{eq:classification-loss}
  \ell_{\rm comp} = \ell_{\rm arc}(\phi_{\rm new}, \omega_{\rm old}),
\end{equation}
where $\omega_{\rm old}$ represents the classifier that was trained on top of the old embedding model.
We also study the contrastive version which constrains the new embeddings to approach the old ones with the same label and keep away from the ones with the different. Here we adopt the supervised contrastive loss~\cite{khosla2020supervised} (specific formulation will be provided in the supplementary material).

Fig.~\ref{fig:compatible-loss-2} displays the trend of ${\cal M}_{\rm BCT}$ using above three types of compatible regularizers. We observe that the gaps between the peak of three curves are distinctly smaller than the difference along the curves itself. What's more, three curves all exhibit an interesting trend that with increasing $\lambda$, ${\cal M}_{\rm BCT}$ first increase and then decrease, which is obviously important for achieving the best backward retrieval performance but not analysed in previous works.
\begin{wrapfigure}{r}{0.5\textwidth}
  \begin{center}
    \includegraphics[width=0.45\textwidth]{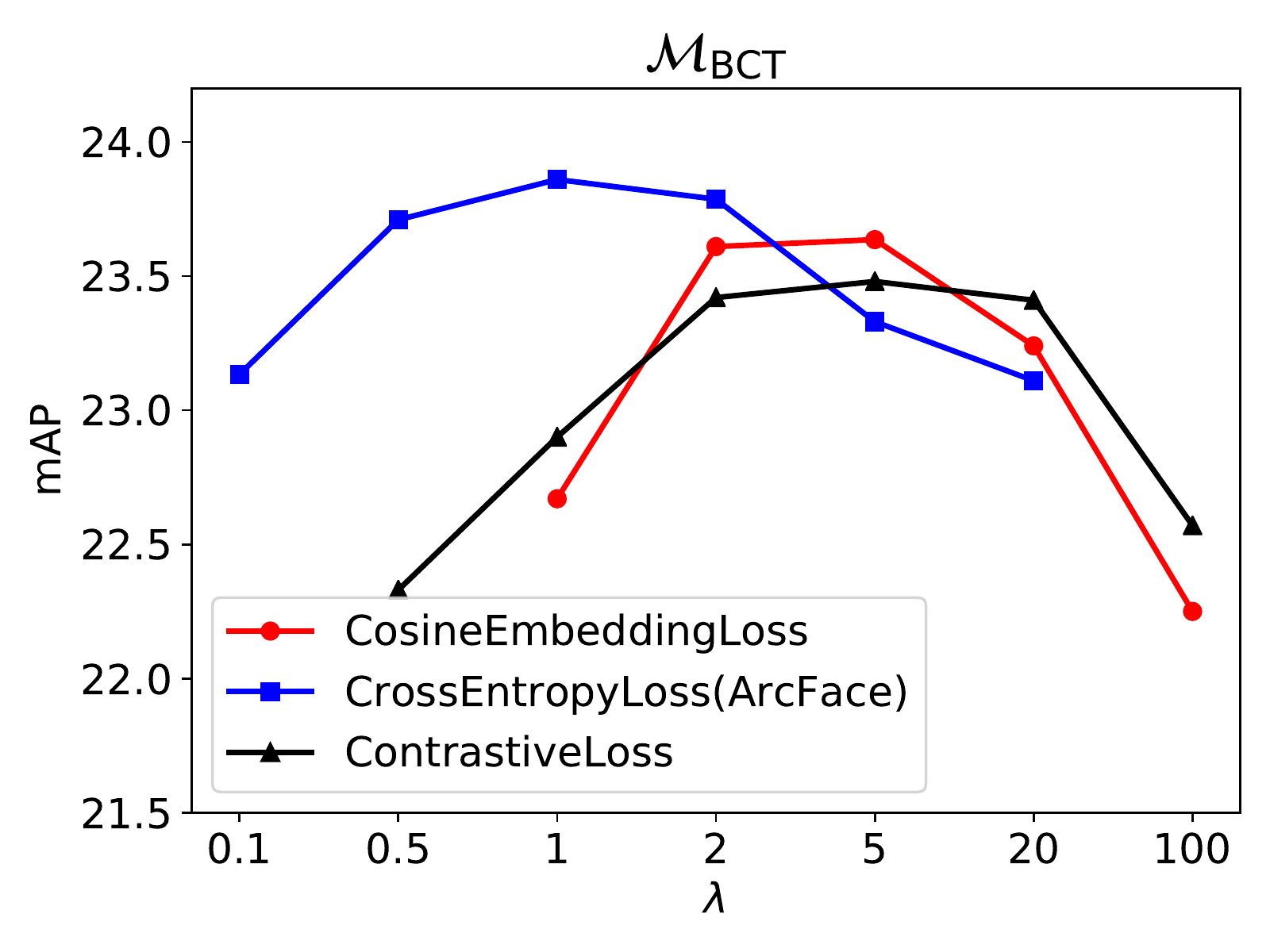}
    \vspace{-5pt}
    \caption{Comparisons for backward-compatible training with three common types of compatible loss under different $\lambda$, when model upgrades from 25\% training data to 100\%.}
    \label{fig:compatible-loss-2}
    \vspace{-16pt}
  \end{center}
  \vspace{-6pt}
\end{wrapfigure}

\begin{figure}[tb]
\centering
\subcaptionbox{Extended~data.}
{\includegraphics[width=0.48\linewidth]{fig/lambda_100data_25data.pdf}\vspace{-0.2cm}}
\hfill
\subcaptionbox{Extended~class.}{\includegraphics[width=0.48\linewidth]{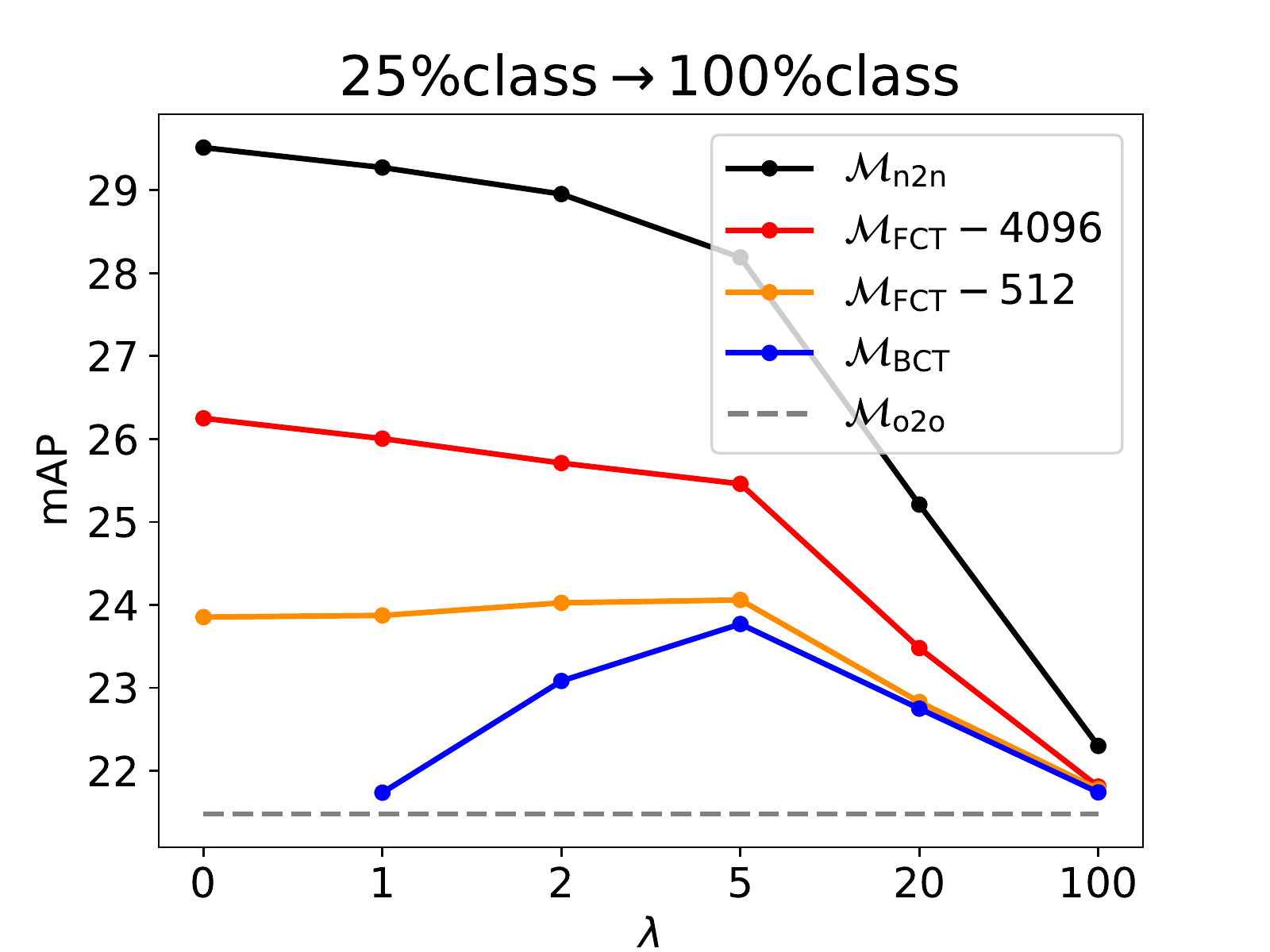}\vspace{-0.2cm}}
\subcaptionbox{Improved~architecture.}{\includegraphics[width=0.48\linewidth]{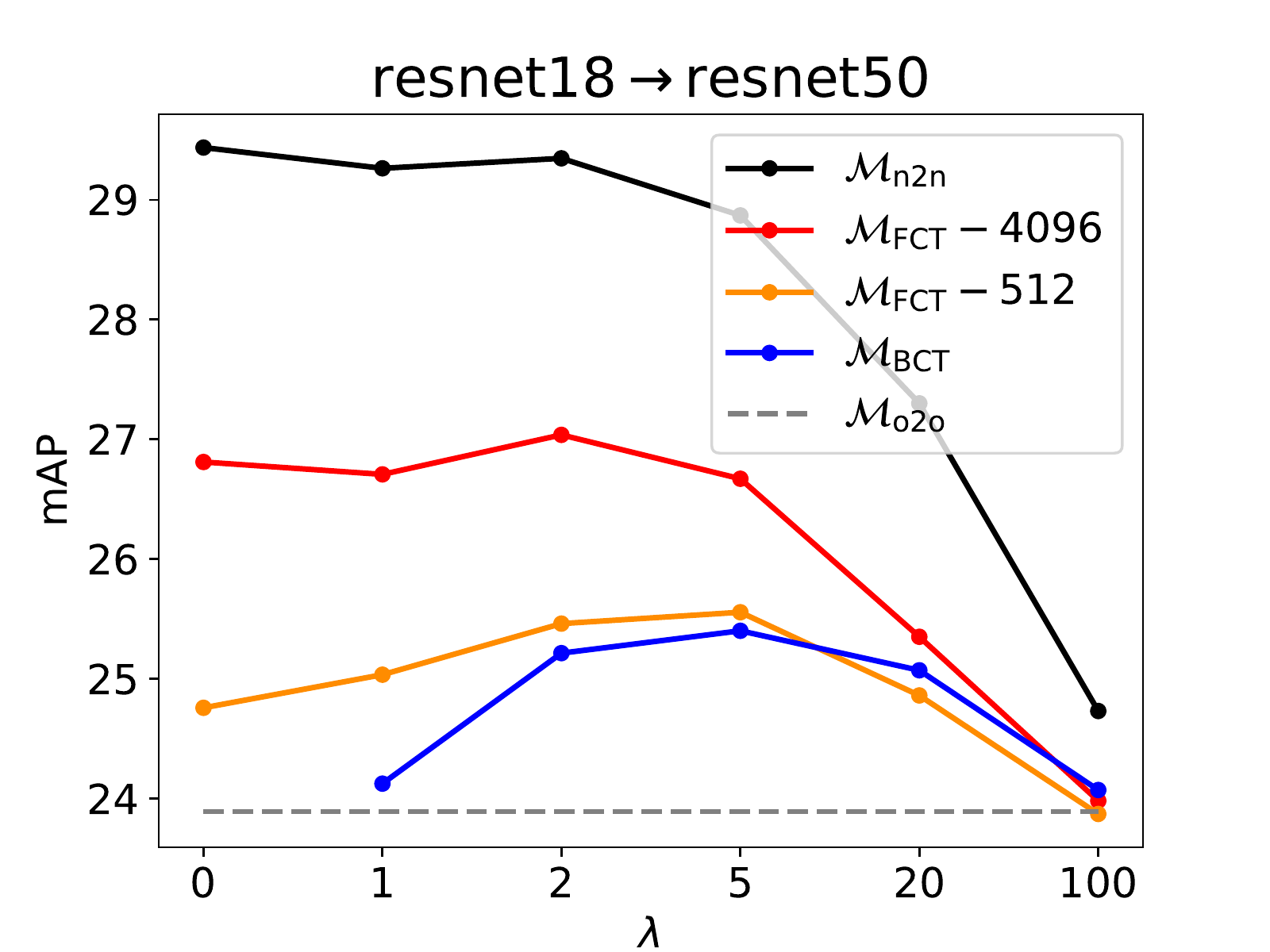}\vspace{-0.2cm}}
\hfill
\subcaptionbox{Improved~loss.}{\includegraphics[width=0.48\linewidth]{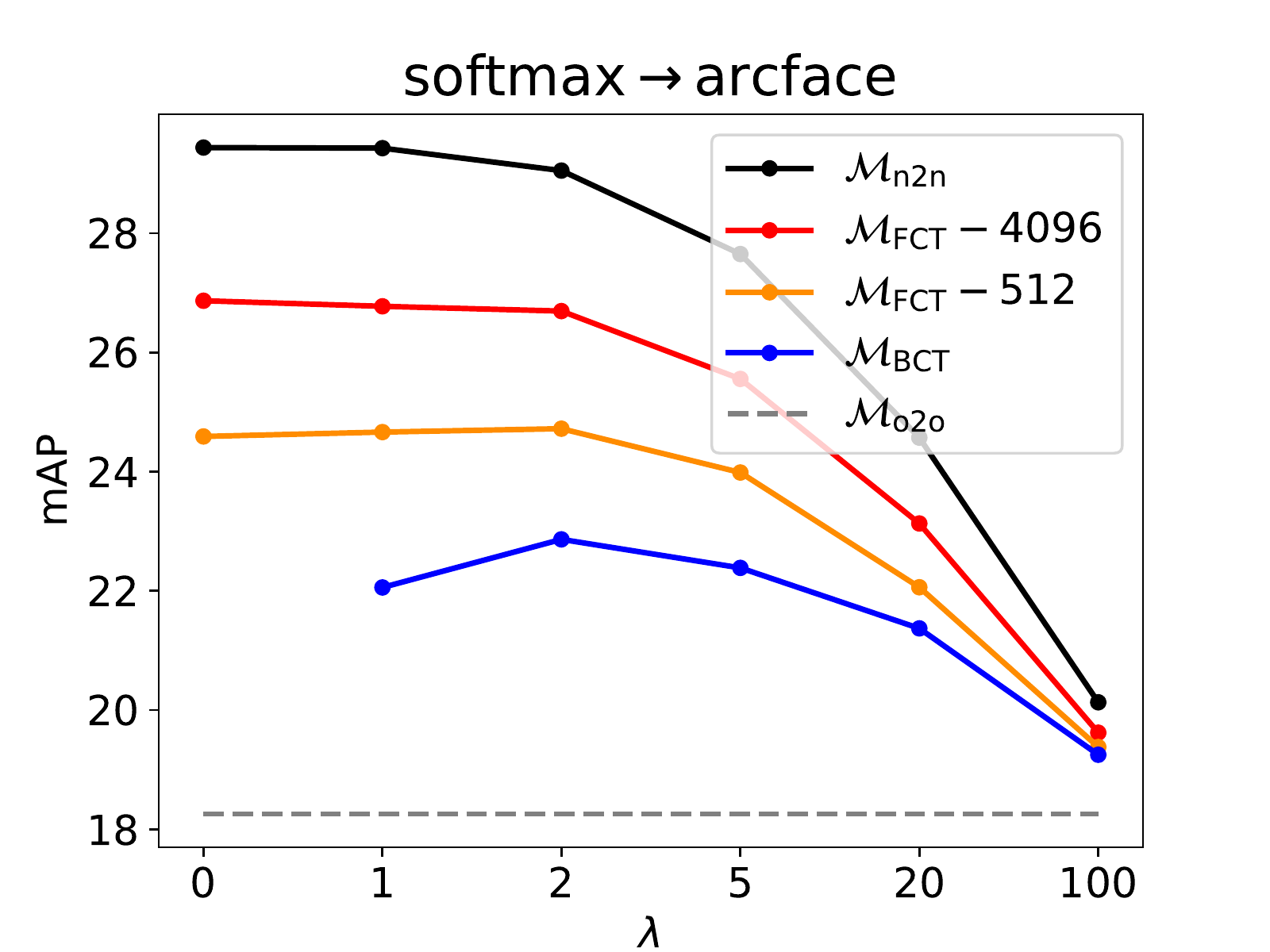}\vspace{-0.2cm}}
\caption{Retrieval performance of BiCT under different $\lambda$ and hidden dimensions of MLP as $\psi$ when model upgrades in the various occasions.}
\label{fig:lambda}
\end{figure}

Therefore, we elaborate the effect of $\lambda$ for both BCT and FCT in the various model upgrade occasions under our BiCT framework.
The results are shown in Fig.~\ref{fig:lambda}.

\paragraph{Effect of $\lambda$ for BCT under BiCT.}
From Fig.~\ref{fig:lambda}, with increasing $\lambda$ from $0$, ${\cal M}_{\rm n2n}$ declines due to the stricter compatible constraint for $\phi_{\rm new}$ towards the weaker $\phi_{\rm old}$. However, ${\cal M}_{\rm BCT}$ shows a peak trend with stronger compatibility.
We figure out that the backward retrieval performance is affected by:
(1) how much performance increase brought by the improvement of query embedding (new embedding capability), and (2) how much performance decline is caused by the discrepancy between the new and old embedding space (compatibility).


Thus, when enlarging $\lambda$ from $0$, the BCT performance first grows (${\cal M}_{\rm BCT}=0$ when $\lambda=0$) due to the prominent promotion of compatibility
than the decline of new embedding capability. 
It keeps rising, exceeding the ${\cal M}_{\rm o2o}$\footnote[5]{${\cal M}_{\rm BCT}$ may not surpass ${\cal M}_{\rm o2o}$ when the improvement of $\phi_{\rm new}$ over $\phi_{\rm old}$ is intrinsically marginal. We skip this special case for clarity.}
until reaching a peak where $\phi_{\rm new}$ is sufficiently compatible to $\phi_{\rm old}$ and simultaneously remains the adequate improvement than $\phi_{\rm old}$.
While in larger $\lambda$, it falls back around ${\cal M}_{\rm o2o}$ as the strengthening enforcement for $\phi_{\rm new}$ to approximate $\phi_{\rm old}$. 
Hence, ${\cal M}_{\rm BCT}$ will always produce a peak trend with increasing $\lambda$.

\paragraph{Effect of $\lambda$ for FCT under BiCT.}
The forward retrieval also depends on the following two aspects:
(1) the new embedding capability of $\phi_{\rm new}$, which is controlled by $\lambda$;
(2) The compatibility between $\phi_{\rm new}$ and $\psi$, which is controlled by both $\lambda$ and the capability of $\psi$ (\ie, the depth or width of MLP). $\lambda$ controls the compatibility between $\phi_{\rm new}$ and $\phi_{\rm old}$ which affects the difficulty of embedding upgrade from $\phi_{\rm old}$ to $\phi_{\rm new}$, while the capability of $\psi$ controls the actual upgrade degree from $\phi_{\rm old}$ to $\phi_{\rm new}$.

Knowing this, with a sufficiently strong $\psi$ that $\psi(\phi_{\rm old})$ can always approximate $\phi_{\rm new}$, ${\cal M}_{\rm FCT}$ will constantly decrease following the decline of new embedding capability in increasing $\lambda$ (the red curve in Fig.~\ref{fig:lambda}). Yet if using a weak $\psi$, the intrinsic compatibility between $\phi_{\rm new}$ and $\phi_{\rm old}$ will occupy a more essential position. Accordingly, ${\cal M}_{\rm FCT}$ ascends at first as the compatible $\phi_{\rm new}$ is closer to the weak $\psi(\phi_{\rm old})$ now (the orange curve in Fig.~\ref{fig:lambda}). 
To take an extreme example, we set $\psi$ as weak as an identity function, \ie, $\psi(x)=x$. In this case, backward performance and forward performance are equivalent (${\cal M}_{\rm FCT}={\cal M}_{\rm BCT}$). Both trends of backward and forward performance start from zero when $\lambda$=0.
After the ascent in the beginning, ${\cal M}_{\rm FCT}$ will descend in larger $\lambda$ due to the dominant decrease of new embedding capability.
Therefore, ${\cal M}_{\rm FCT}$ will either produce a peak or constantly decrease which depends on the capability of $\psi$.


\vspace{-8pt}
\paragraph{Summary.}
To sum up, the BCT performance will always form a peak trend while the FCT performance also constitutes peak with lightweight $\psi$ or continuously decrease (which is slow at first and accelerate later) with burdensome $\psi$. 
We think these analyses not only further demonstrate the two insights we provide in Section~\ref{sec:bict} but also facilitate the in-depth researches in the compatible training community.

\subsection{BiCT in Sequential Model Upgrades}

In the sequential model upgrades occasion, besides recurrently employing the BiCT framework in each generation, here we introduce an effective refinement 
that exploits the momentum update of old features as input for $\psi$. Specifically, we denote $\mathbb{F}_{i}$ as the output feature of $(\psi)_{i}$, where $i$ indicates the $i$-th generation of model upgrade. $\mathbb{F}_{0}$ should be the output feature of $\phi_{\rm old}$ specially. Then, when training $(\psi)_{i}$, $i\ge1$, the input for $(\psi)_{i}$ should be $\mathbb{F}_{i-1}$ normally, but here we propose a momentum update scheme:
\begin{equation} \label{eq:momentum-update}
  \mathrm{input\ for\ } (\psi)_{i} =
  \begin{cases}
  \mathbb{F}_{0}\, , &i=1 \\
  (1-m) \times \mathbb{F}_{i-1} + m \times \mathbb{F}_{i-2}\, , &i\ge2
  \end{cases}
\end{equation}

\noindent
to enrich the information of the input feature for $(\psi)_{i}$ than the mere $\mathbb{F}_{i-1}$.
$m \in [0,1]$ is the momentum hyper-parameter which is set as 0.5 in the following experiments.
Note that we need to save one more old gallery embeddings in this scheme, that is, when we employ $(\psi)_{i}$ to upgrade the gallery features in $(i\!-\!1)$-th generation to $i$-th generation, we need to use $(1-m) \times (\mathbb{F}_{\mathcal{G}})_{i-1} + m \times (\mathbb{F}_{\mathcal{G}})_{i-2}$ instead of $(\mathbb{F}_{\mathcal{G}})_{i-1}$ as input for $(\psi)_{i}$.


We validate the vanilla BiCT and above refinement under the various sequential upgrade occasions and the results are shown in Table~\ref{tab:exp-sequential-upgrades}. We observe that the improvements by FCT are distinct comparing with the mere BCT mechanism especially in the second model generation. It again demonstrates the importance of upgrading the old gallery embeddings after BCT which facilitates both the superior retrieval performance in the current model generation (\ie, ${\cal M}_{\rm FCT}>{\cal M}_{\rm BCT}$) or the future model upgrades (\ie, better ${\cal M}_{\rm BCT}$ of BiCT than mere BCT in the second generation). Besides, the refinement for BiCT brings further promotions across different upgrade settings, which strengthens BiCT for practical applications.

\begin{table}[t]
\renewcommand\arraystretch{1.2}
\footnotesize
\centering
\caption{Sequential model upgrades with BiCT ($\lambda=3$) in three upgrade occasions: extended data, extended class, and improved architecture. `+\textit{mom}' represents that momentum update mechanism (Equation (\ref{eq:momentum-update})) is utilized when training sequential $\psi$.}
\vspace{0.1cm}
\setlength{\tabcolsep}{1.8mm}{	
\begin{tabular}{ccccccc}
	\toprule
	\multirow{2}[3]{*}{Upgrade setting} & \multirow{2}[3]{*}{Method} & \multirow{2}[3]{*}{${\cal M}_{\rm o2o}$} & \multicolumn{2}{c}{1st generation} & \multicolumn{2}{c}{2nd generation} \\
	\cmidrule(r){4-5}
	\cmidrule(r){6-7}
	&&& ${\cal M}_{\rm BCT}$ & ${\cal M}_{\rm FCT}$ & ${\cal M}_{\rm BCT}$ & ${\cal M}_{\rm FCT}$ \\
	\midrule

	\multirow{3}{*}{\makecell[c]{Extended data:\\25\%$\rightarrow$50\%$\rightarrow$100\%}}
	& BCT & \multirow{3}{*}{21.63} & 22.23 & n/a & 23.19 & n/a \\
	& BiCT  && 22.23 & 23.60 & 24.38 & 25.53 \\
	& BiCT+\textit{mom}  && - & - & - & 25.66 \\
	\hline

	\multirow{3}{*}{\makecell[c]{Extended class:\\25\%$\rightarrow$50\%$\rightarrow$100\%}}
	& BCT & \multirow{3}{*}{21.48} & 21.81 & n/a & 23.07 & n/a \\
	& BiCT  && 21.81 & 22.91 & 23.64 & 25.10 \\
	& BiCT+\textit{mom}  && - & - & - & 25.24 \\
	\hline

	\multirow{3}{*}{\makecell[c]{Improved arch:\\resnet18$\rightarrow$resnet34\\$\rightarrow$resnet101}}
	& BCT & \multirow{3}{*}{23.89} & 24.47 & n/a & 25.76 & n/a \\
	& BiCT  && 24.47 & 25.31 & 26.33 & 27.53 \\
	& BiCT+\textit{mom}  && - & - & - & 27.61 \\
	\bottomrule
\end{tabular}
}
\label{tab:exp-sequential-upgrades}
\end{table}

\section{Conclusions and Limitations}
In this paper, we investigate a new and valuable problem, namely privacy-preserving model upgrades, which aims to upgrade the embedding model in the retrieval system without accessing the raw gallery images. We introduce a novel solution, Bidirectional Compatible Training (BiCT), which performs forward-compatible training (FCT) of a feature upgrade module to refresh the old gallery embeddings after the backward-compatible training (BCT) of new model.
The introduction of FCT after BCT properly resolves the dilemma between the demand of upgrading the old gallery embeddings for higher model upgrade performance and the absence of raw gallery images as privacy protection. 
Besides, we demonstrate that the backward-compatible loss weight plays an essential role in both backward and forward-compatible retrieval. The study provides intriguing insights for fully exploring the potential of our BiCT and the research community.

\vspace{6pt}
\noindent\textbf{Limitations.}
Despite that using the forward-compatible module can upgrade the gallery images in an efficient manner during inference,
stacked forward-compatible modules are required for BiCT training after multiple generations, that is, the old model becomes $\psi_N(\cdots(\psi_1(\phi_\text{old}(\cdot))))$, leading to increased computational complexity.
Distilling the stacked models into a single one might be a possible solution; further studies are called for.

%
%
\bibliographystyle{splncs04}
\bibliography{egbib}

\appendix

\section{Implementation Details}

\subsection{Embedding model}

\begin{table}[h]
\renewcommand\arraystretch{1.05}
\footnotesize
\centering
\caption{Training details for the embedding model $\phi_{\rm new}$ and $\phi_{\rm old}$.}
\begin{tabular}{ll}
	\toprule
	config & value \\
	\midrule
	optimizer \hspace{3cm} & SGD \\
	base learning rate & 0.1 \\
	weight decay & 1e-4 \\
	optimizer momentum & 0.9 \\
	\hline
	batch size (per GPU) & 192 \\
	learning schedule & CosineLRScheduler~\cite{loshchilov2016sgdr} \\
	warmup epochs & 1 \\
	training epochs & 30 \\
	initial warmup lr & base lr $\times$ 1e-3 \\
	minimal lr & base lr $\times$ 1e-2 \\
	\hline
	\multirow{2}{*}{training augmentation} & RandomResizedCrop \\
	& \& RandomHorizontalFlip \\
	training image size & 384 $\times$ 384 \\
	inference image size & 640 $\times$ 640 \\
	\hline
	GeM~\cite{radenovic2018fine} param.\ p & 3.0 \\
	Embedding size & 512 \\
	ArcFace~\cite{deng2019arcface} scale & 30.0 \\
	ArcFace~\cite{deng2019arcface} margin & 0.3 \\ 
	\bottomrule
\end{tabular}
\label{tab:model-training-details}
\end{table}

Most of the configurations follow the experience in \cite{yuqi20212nd,ozaki2019large,henkel2020supporting}. Specifically, for the embedding model structure, we adopt the ResNet~\cite{he2016deep} (resnet50 by default) as backbone with the pre-trained weights on ImageNet dataset~\cite{deng2009imagenet}. We substitute the global average pooling layer by Generalized-Mean (GeM) pooling~\cite{radenovic2018fine} with hyper-parameter p=3. A fully-connected (fc) layer is appended in the end which transforms the pooling features into the output embedding (the dimension is set as 512).
We conduct experiments on 8 Tesla V100 and uniformly use the Automatic Mixed Precision (AMP) in both training and inference stages for efficiency.
The complete training details are summarized in Table~\ref{tab:model-training-details}.
Note that for the occasions that needs to adjust the batch size with heavy backbone like resnet101 (whose batch size is set as 128 in our experiments), the base learning rate should be scaled linearly as $lr = 0.1 \times \frac{\#\mathrm{GPUs} \times batchsize}{8\times 192}$.

\subsection{Feature upgrade module}

For the MLP structure as feature upgrade module $\psi$, we adopt [fc-bn-relu]\footnote[6]{bn is short for the Batch Normalization~\cite{ioffe2015batch} layer and relu stands for the ReLU nonlinear activation layer.} as basic block and stack 3 of this with a output fc layer as the final construction. That is, if denoting the hidden dimension as $d_{h}$, the pipeline of MLP we used would be [$d_{old}\!\rightarrow\! d_{h}\!\rightarrow\! d_{h}\!\rightarrow d_{h}\!\rightarrow\! d_{new}$]. 
$\psi$ is trained from scratch and the training details are mostly the same as the ones for embedding model (Table~\ref{tab:model-training-details}) while the different part is listed in Table~\ref{tab:mlp-training-details}.

\begin{table}[h]
\renewcommand\arraystretch{1.05}
\footnotesize
\centering
\caption{Training details for the feature upgrade module $\psi$ where we only list the different settings from the ones in Table~\ref{tab:model-training-details}.}
\begin{tabular}{ll}
	\toprule
	config & value \\
	\midrule
	base learning rate \hspace{3cm} & 1 \\
	batch size (per GPU) & 512 \\
	MLP depth (\#fc-layers) & 4 \\
	MLP width (hidden dimension $d_{h}$) & 4096 \\
	\bottomrule
\end{tabular}
\label{tab:mlp-training-details}
\end{table}

\section{Adaptation of Supervised Contrastive Loss as Compatible Constraint}

We compare three representative versions of compatible regularizer in Section~4.3 of main text. The metric learning type which constrains the new embedding to approach the old features with the same label and keep away from the ones with the different, is adapted from the supervised contrastive loss~\cite{khosla2020supervised} and its specific formulation is listed below:
\begin{equation} \label{eq:contrastive-loss}
  \ell_{\rm comp}(\phi_{\rm new}, \phi_{\rm old}) 
  = \frac{1}{|{\cal D}|}\sum_{i \in \mathcal{D}}
  \frac{1}{\mathcal{P}(i)}
  \sum_{p \in \mathcal{P}(i)} -\log
  \frac{e^{\phi_{\rm new}(i)\phi_{\rm old}(p) / \tau}}
  {\sum_{j \in \mathcal{P}(i) \cup \mathcal{N}(i)} e^{\phi_{\rm new}(i)\phi_{\rm old}(j) / \tau} },
\end{equation}

\noindent
where $\mathcal{P}(i)$ indicates the positive image set for image $i$ and $\mathcal{N}(i)$ indicates the negative set. $\tau$ is the temperature hyper-parameter which is set as $0.1$ in our experiments.

\section{Comparison with Mere FCT Upgrade Mechanism}

It is worth noting that the variant we discuss in the main text which uses mere FCT for model upgrades, is just a special setting ($\lambda=0$) under our proposed BiCT framework. 
Setting $\lambda=0$ means training new model without the backward-compatible constraint whose advantage lies in the potentially better FCT performance when using sufficiently strong $\psi$ as analysed and demonstrated in Section 4.3. 
However, we deprecate this special setting as it will lose the compatibility between the new and old embedding space, which thereby loses the benefits of the immediate deployment of BCT new model and upgrading the old gallery embeddings with $\psi$ on-the-fly.
Instead, we recommend to set a small $\lambda$ as substitute which likewise owns high FCT performance (\eg, from Fig.\ 5 in main text, the FCT performances with $\lambda=1$ are consistently competitive to the ones with $\lambda=0$ in the various upgrade occasions) and simultaneously remains the merits as BiCT in instant deployment of new model and efficient hot-refreshing of gallery embeddings.

\end{document}